\documentclass[11pt]{article}

\usepackage[T1]{fontenc}
\usepackage{lmodern}
\usepackage{microtype}
\usepackage{geometry}
\usepackage{authblk}

\usepackage{amsmath,amssymb,amsfonts}
\usepackage{mathtools}
\usepackage{bm}

\usepackage{amsthm}
\theoremstyle{plain}

\theoremstyle{definition}

\theoremstyle{remark}

\numberwithin{equation}{section}

\usepackage{float}
\usepackage{xurl}

\usepackage{graphicx}
\usepackage{subcaption}
\usepackage{booktabs}
\usepackage{array}
\usepackage{tabularx}
\newcolumntype{Y}{>{\raggedright\arraybackslash}X}

\usepackage[inline,shortlabels]{enumitem}
\setlist{nosep}
\setlist[itemize]{leftmargin=1.2em}
\setlist[enumerate]{leftmargin=1.2em}

\usepackage{algorithm}
\usepackage{algpseudocode}

\usepackage{tikz}
\usetikzlibrary{arrows.meta, positioning, matrix, fit, calc, backgrounds, shadows, shapes.geometric}

\usepackage[numbers,sort&compress]{natbib}

\usepackage[colorlinks=true,linkcolor=magenta,citecolor=blue,urlcolor=blue]{hyperref}
\usepackage[nameinlink,capitalize,noabbrev]{cleveref}

\newcommand{\R}{\mathbb{R}}

\newcommand{\vw}{\bm{w}}

\newcommand{\vr}{\bm{r}}
\newcommand{\vg}{\bm{g}}
\newcommand{\vs}{\bm{s}}

\newcommand{\vv}{\bm{v}} 
\newcommand{\vu}{\bm{u}}

\newcommand{\va}{\bm{a}}

\newcommand{\mI}{\bm{I}}
\newcommand{\mH}{\bm{H}}
\newcommand{\mJ}{\bm{J}}

\newcommand{\mP}{\bm{P}}
\newcommand{\mA}{\bm{A}}
\newcommand{\mD}{\bm{D}}

\newcommand{\norm}[1]{\left\lVert #1 \right\rVert}

\title{
Second-Order, First-Class: A Composable Stack for Curvature-Aware Training
}

\author[1]{Mikalai Korbit}
\author[1]{Mario Zanon}

\affil[1]{IMT School for Advanced Studies Lucca\par Lucca, Italy}

\date{}

\begin{document}

\maketitle

\begin{abstract}
Second-order methods promise improved stability and faster convergence, 
yet they remain underused due to implementation overhead, 
tuning brittleness, and the lack of composable APIs. 
We introduce Somax, a composable Optax-native stack that 
treats curvature-aware training as a single JIT-compiled step governed by a static plan. 
Somax exposes first-class modules -- curvature operators, estimators, 
linear solvers, preconditioners, and damping policies -- behind 
a single step interface
and 
composes with Optax by applying standard gradient transformations 
(e.g., momentum, weight decay, schedules) to the computed direction. 
This design makes typically hidden choices explicit and swappable.
Somax separates planning from execution: 
it derives a static plan (including cadences) 
from module requirements, 
then runs the step through a specialized execution 
path that reuses intermediate results across modules.
We report system-oriented ablations showing that
(i) composition choices materially affect scaling behavior and time-to-accuracy, 
and 
(ii) planning reduces per-step overhead 
relative to unplanned composition with redundant recomputation.
\end{abstract}

\paragraph{Code availability.}
Code corresponding to this paper is available at \url{https://github.com/cor3bit/somax}.

\section{Introduction}
\label{sec:intro}

Second-order optimizers have long promised improved stability and 
better time-to-accuracy in machine learning,
from Hessian-free optimization~\cite{martens2010deep,martens2011learning,kiros2013training}
to Gauss-Newton methods~\cite{ren2019efficient,gargiani2020promise}
and structured preconditioners such as K-FAC and 
Shampoo~\cite{martens2015optimizing,gupta2018shampoo}.
Yet modern training pipelines frequently default to 
Adam~\cite{kingma2014adam} or AdamW~\cite{loshchilov2017decoupled}.
A recurring reason is not a lack of algorithms, 
but rather fragility to implementation choices:
curvature model, linear solver, damping rule, 
and preconditioning can dominate outcomes,
with small changes altering convergence or robustness~\cite{wiesler2013investigations,martens2010deep,martens2011learning}.
In practice, these choices are rarely isolated.
They are embedded in step implementations that 
combine curvature construction, linear solves,
parameter updates, and damping logic,
making it difficult to attribute performance 
to a specific mechanism or 
to reproduce a configuration from a paper description.

We treat this as a systems problem.
A practical second-order method is a multi-stage step pipeline:
(i) construct a curvature snapshot (a linearization state),
(ii) solve a regularized linear system to obtain a direction,
(iii) apply the direction through an update transformation 
(e.g., direction momentum, clipping, weight decay),
(iv) optionally compute post-update signals (e.g., gain ratio for damping control),
and (v) update internal states for the next step.
Multiple components in this pipeline may request 
overlapping computations
(e.g., loss values for trust-region updates and diagnostics, 
or solver statistics for stopping).
Without explicit contracts and execution planning,
implementations either recompute expensive quantities
or couple modules through shared state and implicit assumptions.
Both outcomes hinder experimentation 
and obscure the true cost profile of a step.
Prior work suggests that seemingly small choices inside this pipeline,
including damping updates, solver configuration, estimator cadence,
and warm starts, can dominate stability and wall-clock time-to-accuracy.
A broader related work discussion is provided 
in Appendix~\ref{apx:related}.

We argue that two missing ingredients are \emph{composability} 
and \emph{execution planning}.
Optimizer libraries such as Optax factor training as a chain of
transformations~\cite{deepmind2020jax},
but typical second-order implementations expose functionality 
through solver-centric or framework-specific interfaces
(e.g., KFAC-JAX~\cite{kfac-jax2022github}, JAXopt~\cite{jaxopt_implicit_diff}),
making it difficult to swap a curvature operator 
or damping policy without rewriting the step logic.
Moreover, second-order steps must coordinate 
pre-update and post-update computations:
for instance, trust-region damping needs an actual decrease 
that is consistent with the update that was applied,
which in turn depends on the exact update 
transformation and its internal state.
This demands an execution model 
that retains the linearization 
state long enough to evaluate post-update signals,
while avoiding redundant recomputation.

To address these issues, 
we introduce Somax, 
a composable stack for curvature-aware training in JAX 
with Optax-native post-direction step application.
Somax assembles second-order methods from swappable modules 
with explicit contracts:
curvature operators (e.g., Exact Hessian and Generalized Gauss-Newton),
linear solvers (Conjugate Gradient and direct methods, 
in parameter or row space),
preconditioners (direct diagonal and EMA-based methods),
damping policies (constant, trust-region, and step-norm control),
and estimator-based telemetry 
(diagonal, trace, and spectral summaries
computed from matrix-vector products
without an additional curvature snapshot)~\cite{hutchinson1989stochastic,meyer2021hutch++,liu2023sophia}.
Somax separates static requirements 
(which statistics are needed, and at what cadence)
from a physical execution plan 
(which computations are enabled, cadence-gated, or omitted).
A planner merges module requirements into a static plan that 
fixes the execution lane, metric schema, and cadences,
and a specialized executor performs the step with 
a single linearization and optional post-update policies 
without an additional curvature snapshot.

We evaluate Somax with controlled ablations and system-oriented metrics,
following guidance on optimizer benchmarking~\cite{schneider2019deepobs,schmidt2021descending,moreau2022benchopt}.
In particular, we study how lane and estimator choice, 
solver configuration,
damping, and preconditioning affect scaling behavior
and time-to-accuracy under a fixed step contract.
Lightweight telemetry, such as loss values, gain ratios,
top eigenvalues, and diagonal 
estimates~\cite{yao2020pyhessian,meyer2021hutch++,dangel2019backpack},
is exposed as an explicit cadence-gated systems cost
that can support solver diagnostics and damping control
without being hidden inside optimizer-specific code.

Our contributions are as follows.
\begin{enumerate}
  \item \textbf{End-to-end second-order step API.}
  We provide a step interface that constructs a single step-local curvature state, 
  solves for a direction, 
  applies an Optax transformation to that direction, and updates method and Optax state, with optional post-update signals.

  \item \textbf{Composable module contracts.}
  We expose swappable curvature operators, solvers, damping policies, preconditioners, and estimator-backed telemetry,
  with explicit inputs/outputs and lane-specific 
  constraints enforced at assembly time.

  \item \textbf{Composition-aware execution planning.}
  We merge module requirements into a static step plan 
  that fixes the execution lane, 
  supports reuse of step-local curvature state, 
  and cadence-gates optional computations.

  \item \textbf{Empirical study of module interactions.}
  We report controlled ablations that isolate how solver,
  damping, preconditioning, and execution-lane choices behave
  under a common step interface.

\end{enumerate}

\section{Background: Curvature-Aware Training}
\label{sec:bg}

This section connects the mathematical formulation
of second-order updates with their computational
realization in modern training systems.
We define the learning problem and a first-order baseline,
then introduce curvature-scaled directions 
computed by a damped solve.
Finally, we explain why the same subproblem 
admits three compute regimes (diagonal, parameter-space, and row-space) 
and how Somax makes this choice explicit.

\subsection{Notation}

Scalars are plain (e.g., $a$), 
vectors are bold lowercase (e.g., $\va$), 
and matrices are bold uppercase (e.g., $\mA$).
The Euclidean norm is 
$\norm{\cdot}$ and $\odot$ denotes element-wise products.
The identity matrix is $\mI$.
Indices $i$ and $t$ denote, respectively, data points and iterations.
Curvature operators are denoted by $\mH_t$ and may be exact or approximate; 
Somax accesses $\mH_t$ through matrix-free primitives.
Key symbols are summarized in Table~\ref{tab:notation}.

\begin{table}[H]
\centering
\caption{Notation (excerpt).}
\label{tab:notation}
\begin{tabular}{@{}l p{0.72\columnwidth}@{}}
\toprule
Symbol & Meaning \\
\midrule
$\mathcal D,\,n$ & Dataset and size \\
$\mathcal B_t,\,b$ & Mini-batch at step $t$ and size \\
$\vw \in \R^d$ & Model parameters (weights) \\
$\mathcal L_{\mathcal B_t}(\vw)$ & Mini-batch loss at step $t$ \\
$\vg_t$ & Mini-batch gradient $\nabla \mathcal L_{\mathcal B_t}(\vw_t)$ \\
$\mH_t$ & Curvature operator (exact or approximate) \\
$\lambda_t$ & Damping for the linear subproblem \\
$\vs_t$ & Scaled direction (solution of the subproblem) \\
\bottomrule
\end{tabular}
\end{table}

\subsection{Curvature-Aware Training}

Given data $\mathcal D=\{(x_i,y_i)\}_{i=1}^n$ and a model $f_{\vw}$ 
with per-example loss
$\ell_i(\vw)=\ell(f_{\vw}(x_i),y_i)$, 
we minimize the empirical risk
\begin{align}
\mathcal L_{\mathcal D}(\vw)
\;=\;
\frac{1}{n}\sum_{i=1}^n \ell_i(\vw).
\label{eq:erm}
\end{align}
This empirical risk minimization objective 
is optimized using stochastic mini-batches.
At step $t$, a mini-batch $\mathcal B_t$ of size $b$
yields the batch loss and gradient
\begin{align}
\mathcal L_{\mathcal B_t}(\vw)
\;=\;
\frac{1}{b}\sum_{i\in\mathcal B_t} \ell_i(\vw),
\qquad
\vg_t
\;=\;
\nabla \mathcal L_{\mathcal B_t}(\vw_t).
\label{eq:minibatch}
\end{align}
A first-order update applies a post-processing transform to $\vg_t$ 
(e.g., momentum, clipping, or weight decay) and updates
weights as
\begin{align}
\vw_{t+1}
\;=\;
\vw_t
\;-\;
\alpha_t\,\vu_t,
\label{eq:fo-update}
\end{align}
where $\vu_t$ is the final direction after the chosen first-order transform.

Many optimizers can be viewed as applying a (possibly stateful) linear operator (or preconditioner) to the gradient,
$\vu_t = \mP_t \vg_t$.
First-order methods choose $\mP_t$ to be cheap 
(e.g., diagonal scaling and momentum acceleration),
while second-order methods 
approximate an inverse-curvature operator,
$\mP_t \approx (\mH_t + \lambda_t \mI)^{-1}$,
leading to the damped linear subproblem.

A curvature-aware method modifies the direction computation
by incorporating a local curvature model.
This replaces a purely gradient-based
update with a curvature-scaled direction whose computation is dominated by 
(i) how curvature is represented 
and 
(ii) how the resulting linear system is solved.
This viewpoint explains both the algorithmic diversity 
of second-order methods and the diversity of their implementations.
Different curvature structures and solvers induce 
different dominant primitives and cost profiles.
Table~\ref{tab:som-taxonomy} previews 
the selected methods 
implemented in Somax and used throughout the paper.

A curvature-aware step chooses a (possibly approximate) 
curvature operator $\mH_t$ and defines a scaled direction 
$\vs_t$ via the damped solve
\begin{align}
(\mH_t + \lambda_t \mI)\,\vs_t = \vg_t,
\qquad \lambda_t \ge 0.
\label{eq:scaled-grad}
\end{align}
Damping $\lambda_t$ regularizes the solve and acts 
as a control variable for step selection.
In practice, \eqref{eq:scaled-grad} is often solved 
by conjugate gradient (CG), 
or by preconditioned conjugate gradient (PCG) 
when an explicit preconditioner is available.
When solved inexactly, one typically uses the relative residual criterion
\begin{align}
\norm{(\mH_t+\lambda_t \mI)\vs_t - \vg_t}
\;\le\;
\tau\,\norm{\vg_t}.
\label{eq:inexact_solve}
\end{align}
The applied update may still include the same post-direction transforms 
used by first-order training; 
the key difference is how $\vs_t$ is computed.

\begin{table}[t]
\centering
\caption{Selected second-order methods for large-scale stochastic optimization.}
\label{tab:som-taxonomy}
\setlength{\tabcolsep}{4pt}
\renewcommand{\arraystretch}{1.12}
\begin{tabularx}{\linewidth}{@{}
p{0.20\linewidth}  
p{0.13\linewidth}  
p{0.12\linewidth}  
p{0.16\linewidth}  
Y                  
@{}}
\toprule
\textbf{Method} & \textbf{Curv.} & \textbf{Lane} & \textbf{Solver} & \textbf{Estimation / preconditioning} \\
\midrule
AdaHessian \cite{yao2021adahessian}
& Hessian
& diag.
& explicit
& Hutchinson diag. estimator + diag. EMA preconditioner \\
Sophia-G \cite{liu2023sophia}
& GGN (CE)
& diag.
& explicit
& GNB diag. estimator + diag. EMA preconditioner + clipping \\
Sophia-H \cite{liu2023sophia}
& Hessian
& diag.
& explicit
& Hutchinson diag. estimator + diag. EMA preconditioner + clipping \\
Newton-CG \cite{martens2010deep}
& Hessian
& param.
& CG/PCG (HVP)
& Optional diag. precond.; damping \\
SGN \cite{gargiani2020promise}
& GGN (MSE/CE)
& param.
& CG/PCG (GGN matvec)
& Optional diag. precond.; damping \\
EGN \cite{korbit2025exact}
& GGN (MSE/CE)
& row
& row-Cholesky / row-CG
& Row-space solve with internal damping conversion \\
\bottomrule
\end{tabularx}
\end{table}

Second-order methods in modern ML span a range of curvature structures
and computational strategies.
Diagonal methods retain only per-parameter curvature information
and apply explicit scaling, keeping overhead relatively close
to first-order training; 
examples include AdaHessian~\cite{yao2021adahessian}
and Sophia-style updates~\cite{liu2023sophia}.
Matrix-free methods instead access curvature through 
matrix-vector products 
and solve~\eqref{eq:scaled-grad} iteratively, 
as in Hessian-free optimization~\cite{martens2010deep,martens2011learning,kiros2013training}
and stochastic Gauss-Newton methods~\cite{gargiani2020promise}.
For Gauss-Newton and generalized Gauss-Newton operators,
the same subproblem can also be realized in row space,
leading to a different scaling regime~\cite{ren2019efficient,korbit2024incremental,korbit2025exact}.
Across these families, performance often depends on interacting choices
around the solve, damping, estimation cadence, and preconditioning,
rather than on any single ingredient in isolation.

\subsection{Mapping the Theory to System Design}

Somax targets the compute regimes in Table~\ref{tab:som-taxonomy}
by standardizing two contracts:
(a) curvature access through matrix-free primitives, and
(b) solve execution through one of three lanes.
We describe the operator families used in this paper
and then summarize how each lane realizes the same damped solve
in Equation~\eqref{eq:scaled-grad}.

Two operator families are used throughout the paper.

\emph{Exact Hessian (matrix-free).}
The mini-batch Hessian $\nabla^2 \mathcal L_{\mathcal B_t}(\vw_t)$ 
is applied by Hessian-vector products (HVPs),
computed efficiently via Jacobian-vector product (JVP) 
and vector-Jacobian product (VJP) compositions~\cite{pearlmutter1994fast}.
The raw Hessian can be indefinite, 
damping in \eqref{eq:scaled-grad} is therefore part of the 
definition of the computed direction.

\emph{Generalized Gauss-Newton (GGN).}
For losses of the form $\ell(z,y)$ with $z=f_{\vw}(x)$, 
the GGN takes the form
\begin{align}
\mH_{\mathrm{GGN}}(\vw_t)
=
\frac{1}{b}\sum_{i\in\mathcal B_t}
\mJ_i^\top \mH_{z,i}\,\mJ_i,
\label{eq:ggn}
\end{align}
where $\mJ_i=\nabla_{\vw} f_{\vw}(x_i)$ and
$\mH_{z,i}=\nabla^2_{z}\ell(z,y_i)\big|_{z=f_{\vw}(x_i)}$.
Somax instantiates GGN operators for mean squared error (MSE)
and cross-entropy (CE) objectives and exposes both matvec
and, when available, row-operator primitives.

The same damped subproblem in \eqref{eq:scaled-grad} can be realized through
different execution regimes, which differ in the linear-algebra object
represented explicitly and in the dominant computational primitive
(element-wise scaling, matvecs, or row-space solves).
In Somax, we refer to these execution regimes as \emph{lanes}.

\emph{Diagonal lane.}
The diagonal lane constructs a diagonal approximation
$\mD_t \approx \mH_t + \lambda_t \mI$
(often estimated and smoothed)
and applies an explicit inverse:
\begin{align}
\vs_t = \mD_t^{-1}\vg_t.
\end{align}
This lane covers diagonal-curvature methods
in which curvature is represented as per-parameter scaling.

\emph{Parameter-space solver lane.}
The parameter-space lane defines a matrix-free matvec
$v \mapsto (\mH_t+\lambda_t \mI)v$
and solves~\eqref{eq:inexact_solve} with CG
(or PCG with a supplied preconditioner).
This lane covers Hessian-free / truncated-Newton
and stochastic Gauss-Newton variants 
when the dominant primitive
is a curvature-vector product.

\emph{Row-space solver lane.}
When $\mH_t$ admits a row-operator form, as in GN/GGN methods,
the system can solve in row space and backproject:
\begin{align}
(\mJ\mJ^\top + \mu_t \mI)\,\vv_t = \vr,
\qquad
\vs_t = \mJ^\top \vv_t.
\label{eq:row_system}
\end{align}
This lane exposes a different scaling regime:
the linear system dimension is $b$ rather than $d$.
For mean-reduction objectives, 
we set $\mu_t=b\,\lambda_t$ so that row-space 
regularization matches the parameter-space damping 
convention in \eqref{eq:scaled-grad}.

The remainder of the paper focuses on how this step is realized as a system.
We next describe the Somax planner-executor design
and lane-specialized execution.

\section{Somax: A Composable Stack}\label{sec:somax}

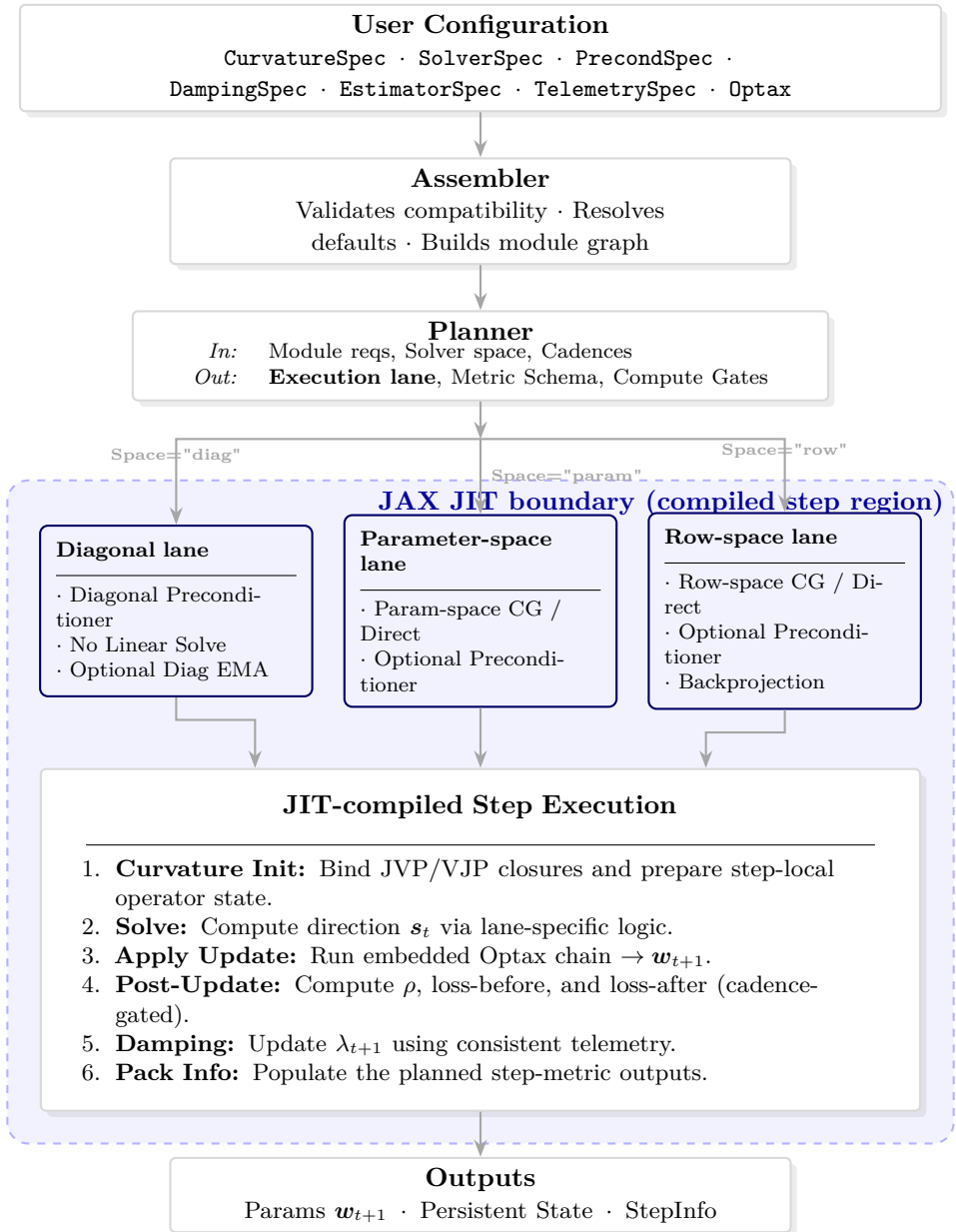
\begin{figure*}[tbp]
\centering
\begin{tikzpicture}[
    font=\sffamily,
    >=Stealth,
    node distance=0.8cm and 0.4cm,
    phasebox/.style={
        rectangle, 
        draw=gray!30, 
        fill=white, 
        thick, 
        rounded corners=2pt, 
        align=center, 
        drop shadow={opacity=0.15},
        font=\small
    },
    lane/.style={
        rectangle, 
        draw=blue!40!black, 
        fill=blue!5, 
        thick, 
        rounded corners=3pt,
        text width=3.2cm, 
        align=left, 
        font=\scriptsize, 
        inner sep=6pt,
        anchor=north
    },
    jitbox/.style={
        rectangle, 
        draw=blue!30, 
        fill=blue!5!white, 
        dashed, 
        thick, 
        rounded corners=8pt,
        inner sep=12pt
    },
    connector/.style={
        ->, thick, gray!70
    },
    process_step/.style={
        rectangle,
        draw=none,
        fill=none,
        text width=10cm,
        align=left,
        font=\small
    }
]


\node[phasebox, text width=12cm] (config) {
    \textbf{User Configuration} \\
    \footnotesize \texttt{CurvatureSpec} $\,\cdot\,$ \texttt{SolverSpec} $\,\cdot\,$ \texttt{PrecondSpec} $\,\cdot\,$ \texttt{DampingSpec}
    $\,\cdot\,$ \texttt{EstimatorSpec} $\,\cdot\,$ \texttt{TelemetrySpec}
    $\,\cdot\,$ \texttt{Optax}
};

\node[phasebox, below=0.6cm of config, text width=8cm] (assembler) {
    \textbf{Assembler} \\
    \footnotesize Validates compatibility $\cdot$ Resolves defaults $\cdot$ Builds module graph
};
\draw[connector] (config) -- (assembler);

\node[phasebox, below=0.6cm of assembler, text width=9cm] (planner) {
    \textbf{Planner} \\
    \scriptsize
    \begin{tabular}{rl}
        \textit{In:} & Module reqs, Solver space, Cadences \\
        \textit{Out:} & \textbf{Execution lane}, Metric Schema, Compute Gates
    \end{tabular}
};
\draw[connector] (assembler) -- (planner);


\node[lane, below=1.5cm of planner] (lane_param) {
    \textbf{Parameter-space lane} \\[-2pt]
    \rule{3.2cm}{0.4pt} \\
    $\cdot$ Param-space CG / Direct \\
    $\cdot$ Optional Preconditioner
};

\node[lane, left=0.4cm of lane_param] (lane_diag) {
    \textbf{Diagonal lane} \\[-2pt]
    \rule{3.2cm}{0.4pt} \\
    $\cdot$ Diagonal Preconditioner \\
    $\cdot$ No Linear Solve \\
    $\cdot$ Optional Diag EMA
};

\node[lane, right=0.4cm of lane_param] (lane_row) {
    \textbf{Row-space lane} \\[-2pt]
    \rule{3.2cm}{0.4pt} \\
    $\cdot$ Row-space CG / Direct \\
    $\cdot$ Optional Preconditioner \\
    $\cdot$ Backprojection
};

\coordinate[below=0.5cm of planner] (split_point);
\draw[connector, thick] (planner.south) -- (split_point);
\draw[connector] (split_point) -| (lane_diag.north) node[pos=0.7, above, font=\tiny\bfseries] {Space="diag"};
\draw[connector] (split_point) -- (lane_param.north) node[midway, right, font=\tiny\bfseries] {Space="param"};
\draw[connector] (split_point) -| (lane_row.north) node[pos=0.7, above, font=\tiny\bfseries] {Space="row"};


\node[phasebox, below=0.8cm of lane_param, text width=11cm, inner sep=10pt] (execution) {
    \textbf{JIT-compiled Step Execution} \\
    \rule{10.5cm}{0.4pt} \\[4pt]
    \begin{minipage}{10.5cm}
    \footnotesize
    \begin{enumerate}
        \itemsep0em 
        \item \textbf{Curvature Init:} Bind JVP/VJP closures and prepare step-local operator state.
        \item \textbf{Solve:} Compute direction $\vs_t$ via lane-specific logic.
        \item \textbf{Apply Update:} Run embedded Optax chain $\to \vw_{t+1}$.
        \item \textbf{Post-Update:} Compute $\rho$, loss-before, and loss-after (cadence-gated).
        \item \textbf{Damping:} Update $\lambda_{t+1}$ using consistent telemetry.
        \item \textbf{Pack Info:} Populate the planned step-metric outputs.
    \end{enumerate}
    \end{minipage}
};

\draw[connector] (lane_diag.south) -- ++(0,-0.3) -| ($(execution.north)+(-3,0)$);
\draw[connector] (lane_param.south) -- (execution.north);
\draw[connector] (lane_row.south) -- ++(0,-0.3) -| ($(execution.north)+(3,0)$);


\node[phasebox, below=0.6cm of execution, text width=8cm] (outputs) {
    \textbf{Outputs} \\
    \footnotesize Params $\vw_{t+1}$ $\,\cdot\,$ Persistent State $\,\cdot\,$ StepInfo
};
\draw[connector] (execution) -- (outputs);

\begin{scope}[on background layer]
    \node[jitbox, fit=(lane_diag) (lane_row) (execution), label={[blue!60!black, anchor=north east, font=\bfseries\small]north east:JAX JIT boundary (compiled step region)}] {};
\end{scope}

\end{tikzpicture}
\caption{
Somax system architecture. 
The assembler and planner resolve a user configuration into a static execution plan. 
This plan records one lane-specific execution path (diagonal, parameter-space, or row-space), the step-metric outputs, and cadence gates; the resulting step is then JIT-compiled. 
The executor manages the full update lifecycle -- from linearization to Optax application and optional post-update control signals -- 
ensuring that post-update metrics are consistent with the applied update.
}
\label{fig:somax-arch}
\end{figure*}

Figure~\ref{fig:somax-arch} summarizes the Somax architecture.
The central design choice 
is to treat curvature-aware optimization
as a \emph{planned step pipeline}
rather than 
as an Optax \texttt{GradientTransformation}.
For first-order methods, returning a transformed direction is often sufficient.
For curvature-aware methods, this abstraction is too weak:
post-update control signals such as actual decrease 
and gain ratio
must be computed from the \emph{applied} update,
which depends on the Optax transform and its internal state.
Somax therefore executes curvature construction, damped solve,
update application, and optional post-update control
within one lane-specialized planned step.

Somax targets single-accelerator JAX execution under JIT.
Its step contract enforces five invariants:
one call to \texttt{step} performs one optimization step;
one curvature snapshot is constructed per step;
the execution lane is fixed at assembly time, with no runtime lane branching;
\texttt{StepInfo} has a fixed structure even when probes are disabled or cadence-gated;
and post-update control signals are computed from the applied update.
These invariants make lane choice, solver configuration, damping,
and estimator cadence configurable under one stable \texttt{init/step} API.
Multi-device execution introduces additional placement and communication issues
and is left to future work.

\subsection{Assembly and Static Planning}
\label{sec:design-planning}

Somax separates method description from method execution.
Users specify curvature, estimator, solver, preconditioning, damping, telemetry,
and Optax post-processing declaratively.
During assembly, defaults are resolved and incompatible combinations are rejected,
so the compiled step contains no runtime compatibility logic.
Assembly also enforces lane-specific constraints,
for example when row-space execution requires row primitives
or disallows incompatible solver features.

The planner then converts the assembled method into a static execution plan.
This plan records:
(i) the execution lane implied by the operator and solver family,
(ii) a fixed \texttt{StepInfo} schema, and
(iii) cadence gates for optional work such as loss-after evaluation,
gain-ratio bundles, and estimator probes.
The planner is deliberately narrow:
it does not search over methods,
but specializes a user-specified configuration into a lane-specific,
JAX-compatible step path.
A component-level contract table is provided in Appendix~\ref{apx:api}.

This plan links the mathematical subproblem in
Section~\ref{sec:bg} to the system behavior studied in Section~\ref{sec:exp}.
By making lane, telemetry, and control paths explicit,
it turns their cost and interactions into measurable properties of one compiled step.

\subsection{Lane-Specialized Compiled Execution}
\label{sec:design-exec}

At runtime, Somax executes exactly one lane-specialized step function
selected by the static plan.
Each step is organized around a single linearization:
the curvature operator is initialized once,
the damped subproblem is solved in the selected lane,
the resulting direction is passed through an internal Optax transformation
to obtain the applied update,
and optional post-update probes are evaluated against that update.
This preserves the semantics required by damping control
while keeping the outer training loop unchanged.

Somax exposes the three execution lanes introduced in
Section~\ref{sec:bg}:
(i) diagonal scaling with no iterative solve,
(ii) parameter-space CG/PCG on matrix-free matvecs, and
(iii) row-space solves, when row primitives are available,
followed by backprojection.
These lanes share the same external step interface,
but differ in their dominant linear-algebra object
and therefore in their scaling behavior.
In particular, the row-space lane is a distinct execution regime,
not a minor implementation variant.

This distinction is what enables the lane-switching study in
Section~\ref{sec:exp:lanes},
which compares different physical realizations of the same damped subproblem under one interface.

\subsection{Planned Telemetry and Cadence-Gated Probes}
\label{sec:design-telemetry}

Somax treats telemetry as part of the step contract rather than as passive logging.
Some signals arise naturally during execution,
such as solver iteration counts or convergence flags.
Others require additional work,
such as loss-after evaluation or gain-ratio bundles 
for damping control.
These optional probes are explicitly planned and cadence-gated.

This design serves two purposes.
First, it preserves semantic consistency:
enabled probes are computed from the same applied update
and, when required, the retained linearization state.
Second, it preserves JAX stability across ablations:
when a probe is disabled or skipped by cadence,
Somax still returns the corresponding \texttt{StepInfo} field
and fills it with a default sentinel.
This keeps the output structure static,
avoids recompilation,
and simplifies downstream logging across method variants.

Telemetry is therefore modeled as optional in-step work
with explicit cadence and explicit cost,
rather than as free instrumentation added after execution.
This makes probe overhead a controlled design variable,
which we isolate in Section~\ref{sec:design-microbench}.

\subsection{Design Validation: Cadence-Gated Probe Overhead}
\label{sec:design-microbench}

Execution planning is a systems claim, not merely a software abstraction.
One concrete consequence is that optional probes should incur cost only when enabled,
and that cost should vary with cadence.
We therefore report a single-device microbenchmark
that isolates cadence-gated post-update probes under a fixed step contract.
This benchmark validates one planner-visible mechanism,
rather than the full benefit of planning.

The benchmark runs on a single NVIDIA RTX A4000 (16GB)
with the JAX GPU backend on a synthetic regression workload with static shapes.
The model is a two-hidden-layer ReLU MLP of width 1024,
with scalar output, input dimension 512, and batch size 256.
We time the JIT-compiled step after warmup
and report steady-state median and p90 step time.


We instantiate a Newton-CG configuration in the parameter-space lane
with constant damping and fixed solver settings,
and vary only the cadence of the post-update \emph{rho} bundle,
which computes actual and predicted decrease and therefore requires
additional post-update work.
All other components are held fixed.
The benchmark sweeps \texttt{rho\_every\_k} over
\texttt{\{-1, 10, 5, 2, 1\}},
where \texttt{-1} disables the probe entirely.

\begin{table}[H]
\centering
\caption{
Microbench: steady-state overhead of cadence-gated rho probes.
Overhead is reported relative to probes disabled.
}
\label{tab:micro_rho_cadence}
\setlength{\tabcolsep}{4pt}
\renewcommand{\arraystretch}{1.1}
\begin{tabular}{lrrr}
\toprule
\rule{0pt}{2.25ex}Setting (\texttt{rho\_every\_k}) & Median (ms) & p90 (ms) & Overhead vs.\ off \\
\midrule
off (\texttt{-1}) & 0.84 & 0.87 & 0.0\% \\
every 10 steps (\texttt{10}) & 0.95 & 1.16 & 12.4\% \\
every 5 steps (\texttt{5}) & 0.96 & 1.33 & 13.8\% \\
every 2 steps (\texttt{2}) & 1.30 & 1.40 & 57.2\% \\
every step (\texttt{1}) & 1.33 & 1.34 & 53.7\% \\
\bottomrule
\end{tabular}
\end{table}

Table~\ref{tab:micro_rho_cadence} shows that post-update probes
are a material systems cost rather than ``free logging''.
Sparse cadences increase steady-state median step time
by about 12-14\% relative to probes disabled,
whereas dense cadences increase it by more than 50\%.
In particular, enabling the rho bundle at every step
raises the median from 0.84\,ms to 1.33\,ms.

This is exactly the contract the planner is meant to expose:
disabled probes incur no cost,
sparse probes incur bounded cost,
and dense probes become a dominant part of the step.
Section~\ref{sec:exp} builds on this execution model
to study lane choice, solver-control interactions,
and estimator cadence under controlled configuration changes.

\section{Experiments}\label{sec:exp}

\subsection{Experimental Setup}

We evaluate Somax as a systems stack 
for curvature-aware training rather than 
as a universal optimizer ranking. 
The experiments comprise three case studies: 
lane-dependent scaling on synthetic regression, 
solver-damping interactions on Fashion-MNIST, 
and estimator-driven diagonal methods 
on CIFAR-10 with ResNet-20.
All experiments use single-accelerator JAX execution under JIT.
Additional implementation and protocol details are provided in Appendix~\ref{apx:exp-details}.

\subsection{Somax Exposes Structure-Aware Execution Regimes}
\label{sec:exp:lanes}

For structured curvature operators, 
execution lane is a first-class systems choice.
Generalized Gauss-Newton (GGN) is a representative case: 
the same curvature family admits either parameter-space solves or structured row-space solves.
Somax exposes this choice explicitly while preserving the same optimizer-facing interface.

This distinction appears concretely 
in classical second-order methods.
Stochastic Gauss-Newton (SGN)~\cite{gargiani2020promise} executes the damped linear solve in parameter space, 
whereas Exact Gauss-Newton (EGN)~\cite{korbit2025exact} transfers the computation to a row-space system 
defined by Jacobian structure.
These methods are therefore a natural pair for isolating the systems consequences of lane choice: 
they share the same curvature family, but their dominant linear-algebra objects scale differently.

We evaluate this effect on a controlled 
synthetic regression workload 
designed to vary the two quantities that most 
directly determine the relative cost of the two lanes:
parameter dimension and batch size.
We compare two matched second-order configurations that differ only in execution lane, 
namely a parameter-space CG configuration 
and a row-space CG configuration.
The loss, outer update semantics, learning rate, damping policy, and inner-solver budget are held fixed.

\begin{figure}[t]
\centering
\includegraphics[width=\textwidth]{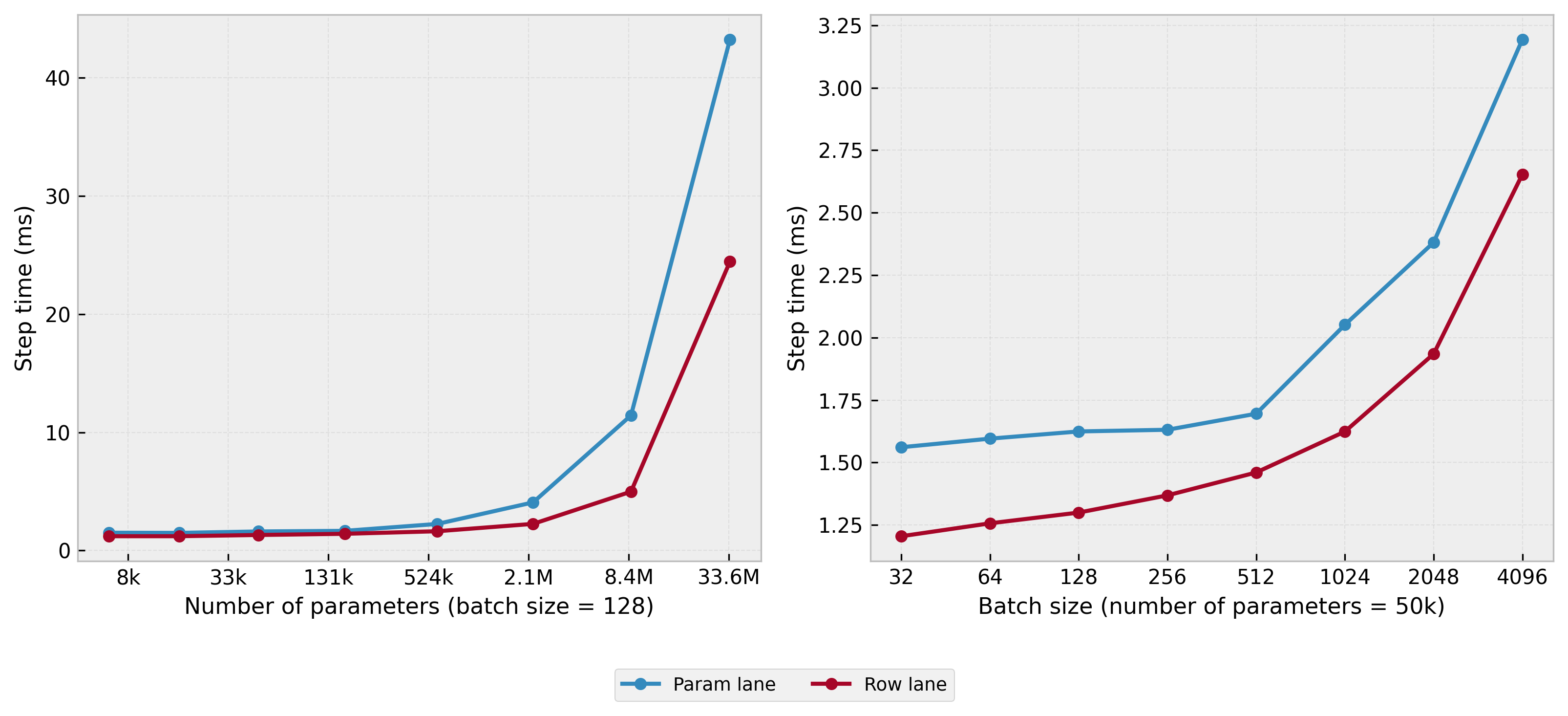}
\caption{
Steady-state step time for matched parameter-space and row-space configurations under a shared training interface.
Left: increasing model size at fixed batch size. Right: increasing batch size at fixed model size.
}
\label{fig:exp:lane-regimes}
\end{figure}

Figure~\ref{fig:exp:lane-regimes} reports 
steady-state step time under two sweeps.
In the left panel, we fix batch size 
and increase model size, thereby increasing parameter dimension while keeping the row-space dimension fixed.
In the right panel, we fix model size and increase batch size, thereby increasing the row-space dimension while keeping parameter dimension fixed.
The resulting curves expose distinct scaling signatures.
As model size grows, the parameter lane becomes substantially more expensive, 
while the row lane grows more gradually.
As batch size grows, both lanes slow down, 
but the row lane retains a consistent advantage throughout the tested range.

The key observation is not that one regime universally dominates.
Rather, when curvature structure can be exploited,
Somax turns that opportunity into an explicit execution choice with its own scaling law.
Changing the execution regime changes the dominant 
linear-algebra object and therefore the cost regime 
of the second-order step, 
even though the outer \texttt{init/step} interface remains unchanged.

\subsection{Interactions Between Modules Shape Performance}
\label{sec:exp:interactions}

\begin{table}[t]
\centering
\caption{Interactions between solver modules shape 
performance on Fashion-MNIST.}
\label{tab:solver-comparison}
\small
\begin{tabular}{l l l l r r}
\toprule
\textbf{Solver} & \textbf{Precond.} & \textbf{Damping} & \textbf{CG budget} & \textbf{Time to 85\% (s)} & \textbf{Final acc (\%)} \\
\midrule
SGD  & --     & --           & --     & 1.60 & 90.58 \\
Adam & --     & --           & --     & 1.23 & 90.39 \\
\midrule
SGN  & --     & const        & light  & 3.52 & 90.69 \\
SGN  & --     & const        & heavy  & 4.79 & 90.09 \\
SGN  & --     & trust region & light  & 7.33 & 89.69 \\
SGN  & --     & trust region & heavy  & 9.13 & 91.14 \\
\midrule
SGN  & sq grad & const        & light  & 3.54 & 90.81 \\
SGN  & sq grad & const        & heavy  & 5.48 & 90.27 \\
SGN  & sq grad & trust region & light  & --   & 19.46 \\
SGN  & sq grad & trust region & heavy  & --   & 9.03  \\
\bottomrule
\end{tabular}
\end{table}

Prior work on practical second-order optimization 
has repeatedly shown that performance is governed 
not only by the curvature model itself, 
but also by the interaction between linear solves, 
damping, and preconditioning~\cite{martens2010deep,wiesler2013investigations}.
Somax is designed around this observation: 
rather than treating a second-order method 
as a monolithic optimizer, 
it exposes the main control components as composable modules under a shared step interface.

This case study fixes the parameter-space SGN setting 
on Fashion-MNIST and varies three modules: 
the damping policy, the CG budget, and 
the use of diagonal preconditioning.
All runs share the same dataset, model, batch size, and training loop.
For each configuration, 
we perform a learning rate search, 
selecting the best rate, 
and then report results averaged over 5 seeds.
We include SGD and Adam for context, 
but the primary purpose of the experiment is to 
compare second-order module combinations under fixed lane and curvature choices.
We use two predefined CG budgets:
the \emph{light} budget uses a small iteration cap and loose tolerance, while the 
\emph{heavy} budget uses a larger cap and tighter tolerance.
Exact model and protocol details, 
including the model architecture 
and the exact definitions of the light and heavy CG budgets, 
are given in Appendix~\ref{apx:exp-details}.

Table~\ref{tab:solver-comparison} summarizes the resulting interaction study.
Several interaction effects are immediately visible.
First, damping strongly changes the usefulness of a given CG budget.
With constant damping, 
the light budget reaches the target faster 
than the heavy budget while achieving similar final accuracy.
With trust-region damping,
both configurations are slower, 
but the heavy budget attains the best final accuracy among the SGN variants.
Second, increased solve effort does not uniformly improve performance.
Heavier CG budgets consistently increase wall-clock time-to-target, 
and under constant damping they do so without improving final accuracy.
Third, diagonal preconditioning is not uniformly beneficial in this setting.
With constant damping it behaves similarly to the non-preconditioned baseline, with nearly identical time-to-target and final accuracy.
Under trust-region damping, 
however, diagonal preconditioning performs poorly: 
the light and heavy configurations reach only 19.46\% 
and 9.03\% final accuracy, respectively. 
This suggests that module compatibility cannot be assumed,
as the chosen diagonal preconditioner interacts unfavorably 
with trust-region acceptance dynamics.

We show that practical second-order behavior within 
a fixed lane emerges from module interactions 
rather than isolated knobs. 
The goal is not to identify a universally best method, 
but to show that changing a small set of modules can materially affect performance.
Somax makes these interactions efficient, explicit, 
swappable, and measurable under a shared execution contract, 
turning solver design from ad hoc optimizer rewriting 
to auditable configuration.

\subsection{Composability Enables New Optimizer Regimes}

\begin{figure}[t]
\centering
\includegraphics[width=0.65\linewidth]{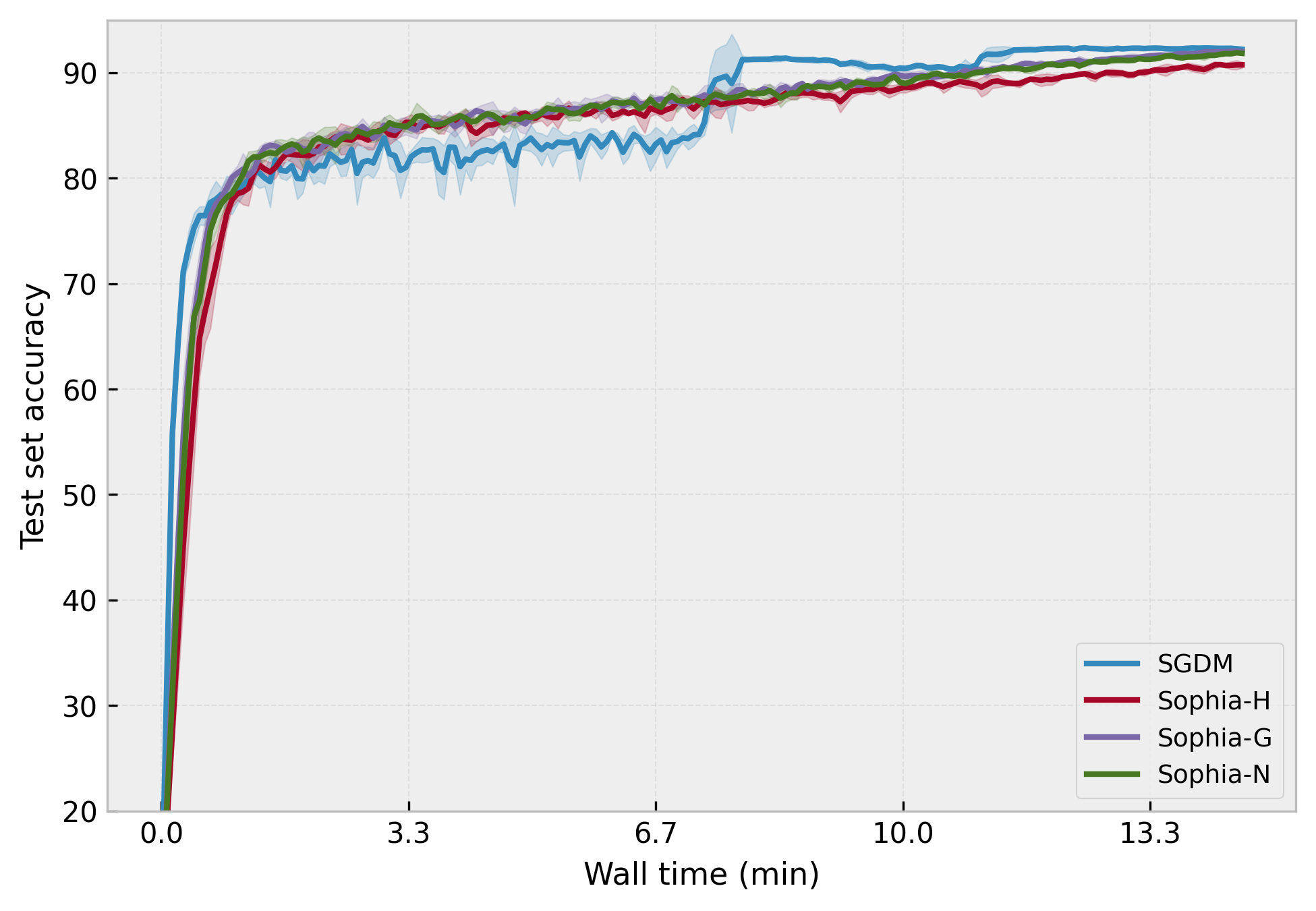}
\caption{
Wall-clock performance comparison of 
SGDM and Sophia-style optimizer regimes on 
CIFAR-10 with ResNet-20.
Thick lines show mean test accuracy across 5 seeds; shaded bands indicate $\pm 1$ standard deviation.
}
\label{fig:exp:sophia}
\end{figure}

Somax treats second-order optimizers as 
compositions of modular components rather 
than fixed algorithms.
Under this view, Sophia-style methods can be expressed as particular choices of curvature operator, 
diagonal estimator, refresh cadence, and outer update logic.
For example, Sophia-H corresponds to an Exact Hessian curvature operator combined with Hutchinson diagonal estimation and diagonal EMA preconditioning, 
while Sophia-G replaces the curvature-estimation pair with GGN curvature and the Gauss-Newton-Bartlett estimator~\cite{liu2023sophia}.
This factorization exposes a broader design space in which related diagonal second-order regimes can be assembled under the same training interface.

To illustrate this design space, 
we construct a new Sophia-style regime on 
CIFAR-10 with ResNet-20 by combining GGN curvature 
with Hutchinson diagonal estimation.
We denote this configuration as Sophia-N.
The point is not to claim algorithmic novelty in isolation, 
but to show that Somax makes such recombinations straightforward 
to express and evaluate within a shared training pipeline.
All methods use the same model and surrounding training loop, 
with SGDM included as a first-order reference.

Figure~\ref{fig:exp:sophia} 
reports test accuracy against wall-clock time. 
SGDM achieves the strongest final performance within the common wall-clock budget, 
reaching $92.21 \pm 0.15$ test accuracy.
Among the Sophia-style methods, Sophia-G and Sophia-N perform similarly, reaching $91.99 \pm 0.33$ and $91.83 \pm 0.13$, respectively, while Sophia-H trails at $90.74 \pm 0.29$.
Sophia-N converges competitively with established Sophia variants, 
showing that modular recombination 
can produce new configurations 
without requiring monolithic optimizer implementations.
This is the main systems point of the experiment: once the optimizer is factorized into interchangeable modules, 
new optimizer regimes become concrete, 
testable assemblies 
rather than one-off optimizer implementations.

\section{Conclusions and Future Work}
\label{sec:conclusions}

We introduced Somax,
a composable Optax-native stack
that makes curvature-aware training a planned, compiled step pipeline.
Each step constructs a single step-local curvature state,
runs a lane-specialized solve (diagonal, parameter-space, or row-space),
applies the direction through an embedded Optax update transformation,
and updates auxiliary method state.
A planner records the execution lane,
the fixed output structure for step metrics,
and cadences for optional computations,
yielding a stable step interface for controlled second-order experiments.

Our experiments support three design claims.
First, execution lane is a first-class architectural choice:
parameter-space and row-space realizations have different cost models
and scaling regimes under the same step contract.
Second, solver behavior is coupled to damping and preconditioning;
Somax makes this interaction measurable by computing post-update control signals
from the applied update.
Third, the same contract can enable new optimizer regimes through
modular recombination.

\paragraph{Future work.}
Somax is designed to admit further curvature-aware 
optimizers as contract-preserving extensions.
Natural next steps include structured preconditioners, 
such as Kronecker-factored~\cite{martens2015optimizing} or 
tensor-axis methods~\cite{gupta2018shampoo}. 
Quasi-Newton methods~\cite{berahas2016multi,byrd2016stochastic,wills2021stochastic} 
can be integrated as memory-bounded preconditioners, 
including diagonal-lane variants.
Randomized and sketch-based curvature
approximations~\cite{mahoney2011randomized,pilanci2017newton} 
offer another axis, enabling rank-controlled operator primitives 
while preserving the same step interface.

The main systems extension is multi-device execution.
Distributed curvature-aware training introduces additional concerns,
including parameter and optimizer-state sharding,
communication-aware lane placement,
and collective operations for curvature statistics.

\bibliographystyle{plainnat}
\bibliography{references}

@article{kingma2014adam,
  title={Adam: A method for stochastic optimization},
  author={Kingma, Diederik P and Ba, Jimmy},
  journal={arXiv preprint arXiv:1412.6980},
  year={2014}
}

@article{loshchilov2017decoupled,
  title={Decoupled weight decay regularization},
  author={Loshchilov, Ilya and Hutter, Frank},
  journal={arXiv preprint arXiv:1711.05101},
  year={2017}
}

@article{gargiani2020promise,
  title={On the Promise of the Stochastic Generalized {G}auss-{N}ewton Method for Training {DNNs}},
  author={Gargiani, Matilde and Zanelli, Andrea and Diehl, Moritz and Hutter, Frank},
  journal={arXiv preprint arXiv:2006.02409},
  year={2020}
}

@article{ren2019efficient,
  title={Efficient subsampled Gauss-Newton and natural gradient methods for training neural networks},
  author={Ren, Yi and Goldfarb, Donald},
  journal={arXiv preprint arXiv:1906.02353},
  year={2019}
}

@inproceedings{martens2010deep,
  title={Deep learning via {H}essian-free optimization.},
  author={Martens, James and others},
  booktitle={ICML},
  volume={27},
  pages={735--742},
  year={2010}
}

@inproceedings{martens2011learning,
  title={Learning recurrent neural networks with {H}essian-free optimization},
  author={Martens, James and Sutskever, Ilya},
  booktitle={Proceedings of the 28th international conference on machine learning (ICML-11)},
  pages={1033--1040},
  year={2011}
}

@inproceedings{wiesler2013investigations,
  title={Investigations on {H}essian-free optimization for cross-entropy training of deep neural networks.},
  author={Wiesler, Simon and Li, Jinyu and Xue, Jian},
  booktitle={Interspeech},
  pages={3317--3321},
  year={2013}
}

@inproceedings{martens2015optimizing,
  title={Optimizing neural networks with kronecker-factored approximate curvature},
  author={Martens, James and Grosse, Roger},
  booktitle={International conference on machine learning},
  pages={2408--2417},
  year={2015},
  organization={PMLR}
}

@article{kiros2013training,
  title={Training neural networks with stochastic {H}essian-free optimization},
  author={Kiros, Ryan},
  journal={arXiv preprint arXiv:1301.3641},
  year={2013}
}

@article{papyan2018full,
  title={The full spectrum of deep net {H}essians at scale: Dynamics with sample size},
  author={Papyan, Vardan},
  journal={arXiv preprint arXiv:1811.07062},
  year={2018}
}

@article{byrd2016stochastic,
  title={A stochastic quasi-{N}ewton method for large-scale optimization},
  author={Byrd, Richard H and Hansen, Samantha L and Nocedal, Jorge and Singer, Yoram},
  journal={SIAM Journal on Optimization},
  volume={26},
  number={2},
  pages={1008--1031},
  year={2016},
  publisher={SIAM}
}

@article{berahas2016multi,
  title={A multi-batch {L-BFGS} method for machine learning},
  author={Berahas, Albert S and Nocedal, Jorge and Tak{\'a}c, Martin},
  journal={Advances in Neural Information Processing Systems},
  volume={29},
  year={2016}
}

@article{wills2021stochastic,
  title={Stochastic quasi-{N}ewton with line-search regularisation},
  author={Wills, Adrian G and Sch{\"o}n, Thomas B},
  journal={Automatica},
  volume={127},
  pages={109503},
  year={2021},
  publisher={Elsevier}
}

@article{liu2023sophia,
  title={Sophia: A Scalable Stochastic Second-order Optimizer for Language Model Pre-training},
  author={Liu, Hong and Li, Zhiyuan and Hall, David and Liang, Percy and Ma, Tengyu},
  journal={arXiv preprint arXiv:2305.14342},
  year={2023}
}

@inproceedings{yao2021adahessian,
  title={Adahessian: An adaptive second order optimizer for machine learning},
  author={Yao, Zhewei and Gholami, Amir and Shen, Sheng and Mustafa, Mustafa and Keutzer, Kurt and Mahoney, Michael},
  booktitle={proceedings of the AAAI conference on artificial intelligence},
  volume={35},
  pages={10665--10673},
  year={2021}
}

@article{jaxopt_implicit_diff,
  title={Efficient and Modular Implicit Differentiation},
  author={Blondel, Mathieu and Berthet, Quentin and Cuturi, Marco and Frostig, Roy 
    and Hoyer, Stephan and Llinares-L{\'o}pez, Felipe and Pedregosa, Fabian 
    and Vert, Jean-Philippe},
  journal={arXiv preprint arXiv:2105.15183},
  year={2021}
}

@misc{deepmind2020jax,
  title = {The {D}eep{M}ind {JAX} {E}cosystem},
  author = {DeepMind and Babuschkin, Igor and Baumli, Kate and Bell, Alison and Bhupatiraju, Surya and Bruce, Jake and Buchlovsky, Peter and Budden, David and Cai, Trevor and Clark, Aidan and Danihelka, Ivo and Dedieu, Antoine and Fantacci, Claudio and Godwin, Jonathan and Jones, Chris and Hemsley, Ross and Hennigan, Tom and Hessel, Matteo and Hou, Shaobo and Kapturowski, Steven and Keck, Thomas and Kemaev, Iurii and King, Michael and Kunesch, Markus and Martens, Lena and Merzic, Hamza and Mikulik, Vladimir and Norman, Tamara and Papamakarios, George and Quan, John and Ring, Roman and Ruiz, Francisco and Sanchez, Alvaro and Sartran, Laurent and Schneider, Rosalia and Sezener, Eren and Spencer, Stephen and Srinivasan, Srivatsan and Stanojevi\'{c}, Milo\v{s} and Stokowiec, Wojciech and Wang, Luyu and Zhou, Guangyao and Viola, Fabio},
  url = {http://github.com/google-deepmind},
  year = {2020},
}

@article{rader2023lineax,
  title={Lineax: unified linear solves and linear least-squares in JAX and Equinox},
  author={Rader, Jason and Lyons, Terry and Kidger, Patrick},
  journal={arXiv preprint arXiv:2311.17283},
  year={2023}
}

@article{schneider2021cockpit,
  title={Cockpit: A practical debugging tool for the training of deep neural networks},
  author={Schneider, Frank and Dangel, Felix and Hennig, Philipp},
  journal={Advances in Neural Information Processing Systems},
  volume={34},
  pages={20825--20837},
  year={2021}
}

@article{korbit2025exact,
  title={Exact gauss-newton optimization for training deep neural networks},
  author={Korbit, Mikalai and Adeoye, Adeyemi D and Bemporad, Alberto and Zanon, Mario},
  journal={Neurocomputing},
  pages={131738},
  year={2025},
  publisher={Elsevier}
}

@article{korbit2024incremental,
  title={Incremental Gauss-Newton Descent for Machine Learning},
  author={Korbit, Mikalai and Zanon, Mario},
  journal={arXiv preprint arXiv:2408.05560},
  year={2024}
}

@inproceedings{gupta2018shampoo,
  title={Shampoo: Preconditioned stochastic tensor optimization},
  author={Gupta, Vineet and Koren, Tomer and Singer, Yoram},
  booktitle={International Conference on Machine Learning},
  pages={1842--1850},
  year={2018},
  organization={PMLR}
}

@inproceedings{osawa2019large,
  title={Large-scale distributed second-order optimization using kronecker-factored approximate curvature for deep convolutional neural networks},
  author={Osawa, Kazuki and Tsuji, Yohei and Ueno, Yuichiro and Naruse, Akira and Yokota, Rio and Matsuoka, Satoshi},
  booktitle={Proceedings of the IEEE/CVF Conference on Computer Vision and Pattern Recognition},
  pages={12359--12367},
  year={2019}
}

@article{osawa2020scalable,
  title={Scalable and practical natural gradient for large-scale deep learning},
  author={Osawa, Kazuki and Tsuji, Yohei and Ueno, Yuichiro and Naruse, Akira and Foo, Chuan-Sheng and Yokota, Rio},
  journal={IEEE Transactions on Pattern Analysis and Machine Intelligence},
  volume={44},
  number={1},
  pages={404--415},
  year={2020},
  publisher={IEEE}
}

@article{pauloski2022deep,
  title={Deep neural network training with distributed K-FAC},
  author={Pauloski, J Gregory and Huang, Lei and Xu, Weijia and Chard, Kyle and Foster, Ian T and Zhang, Zhao},
  journal={IEEE Transactions on Parallel and Distributed Systems},
  volume={33},
  number={12},
  pages={3616--3627},
  year={2022},
  publisher={IEEE}
}

@article{zhang2022scalable,
  title={Scalable k-fac training for deep neural networks with distributed preconditioning},
  author={Zhang, Lin and Shi, Shaohuai and Wang, Wei and Li, Bo},
  journal={IEEE Transactions on Cloud Computing},
  volume={11},
  number={3},
  pages={2365--2378},
  year={2022},
  publisher={IEEE}
}

@article{vyas2024soap,
  title={Soap: Improving and stabilizing shampoo using adam},
  author={Vyas, Nikhil and Morwani, Depen and Zhao, Rosie and Kwun, Mujin and Shapira, Itai and Brandfonbrener, David and Janson, Lucas and Kakade, Sham},
  journal={arXiv preprint arXiv:2409.11321},
  year={2024}
}

@inproceedings{yao2020pyhessian,
  title={Pyhessian: Neural networks through the lens of the hessian},
  author={Yao, Zhewei and Gholami, Amir and Keutzer, Kurt and Mahoney, Michael W},
  booktitle={2020 IEEE international conference on big data (Big data)},
  pages={581--590},
  year={2020},
  organization={IEEE}
}

@article{dangel2019backpack,
  title={Backpack: Packing more into backprop},
  author={Dangel, Felix and Kunstner, Frederik and Hennig, Philipp},
  journal={arXiv preprint arXiv:1912.10985},
  year={2019}
}

@inproceedings{meyer2021hutchpp,
  title={Hutch++: Optimal stochastic trace estimation},
  author={Meyer, Raphael A and Musco, Cameron and Musco, Christopher and Woodruff, David P},
  booktitle={Symposium on Simplicity in Algorithms (SOSA)},
  pages={142--155},
  year={2021},
  organization={SIAM}
}

@inproceedings{duvvuri2024caspr,
  title={CASPR: Combining axes preconditioners through Kronecker approximation for deep learning},
  author={Duvvuri, Sai Surya and Devvrit, Fnu and Anil, Rohan and Hsieh, Cho-Jui and Dhillon, Inderjit},
  booktitle={Forty-first International Conference on Machine Learning},
  year={2024}
}

@article{schneider2019deepobs,
  title={Deepobs: A deep learning optimizer benchmark suite},
  author={Schneider, Frank and Balles, Lukas and Hennig, Philipp},
  journal={arXiv preprint arXiv:1903.05499},
  year={2019}
}

@inproceedings{schmidt2021descending,
  title={Descending through a crowded valley-benchmarking deep learning optimizers},
  author={Schmidt, Robin M and Schneider, Frank and Hennig, Philipp},
  booktitle={International Conference on Machine Learning},
  pages={9367--9376},
  year={2021},
  organization={PMLR}
}

@article{moreau2022benchopt,
  title={Benchopt: Reproducible, efficient and collaborative optimization benchmarks},
  author={Moreau, Thomas and Massias, Mathurin and Gramfort, Alexandre and Ablin, Pierre and Bannier, Pierre-Antoine and Charlier, Benjamin and Dagr{\'e}ou, Mathieu and Dupre la Tour, Tom and Durif, Ghislain and Dantas, Cassio F and others},
  journal={Advances in Neural Information Processing Systems},
  volume={35},
  pages={25404--25421},
  year={2022}
}

@misc{kfac-jax2022github,
  author = {Aleksandar Botev and James Martens},
  title = {{KFAC-JAX}},
  url = {https://github.com/google-deepmind/kfac-jax},
  version = {0.0.2},
  year = {2022},
}

@article{optimistix2024,
    title={Optimistix: modular optimisation in JAX and Equinox},
    author={Jason Rader and Terry Lyons and Patrick Kidger},
    journal={arXiv:2402.09983},
    year={2024},
}

@article{hutchinson1989stochastic,
  title={A stochastic estimator of the trace of the influence matrix for Laplacian smoothing splines},
  author={Hutchinson, Michael F},
  journal={Communications in Statistics-Simulation and Computation},
  volume={18},
  number={3},
  pages={1059--1076},
  year={1989},
  publisher={Taylor \& Francis}
}

@inproceedings{meyer2021hutch++,
  title={Hutch++: Optimal stochastic trace estimation},
  author={Meyer, Raphael A and Musco, Cameron and Musco, Christopher and Woodruff, David P},
  booktitle={Symposium on Simplicity in Algorithms (SOSA)},
  pages={142--155},
  year={2021},
  organization={SIAM}
}

@article{pearlmutter1994fast,
  title={Fast exact multiplication by the Hessian},
  author={Pearlmutter, Barak A},
  journal={Neural computation},
  volume={6},
  number={1},
  pages={147--160},
  year={1994},
  publisher={MIT Press}
}

@article{mahoney2011randomized,
  title={Randomized algorithms for matrices and data},
  author={Mahoney, Michael W and others},
  journal={Foundations and Trends{\textregistered} in Machine Learning},
  volume={3},
  number={2},
  pages={123--224},
  year={2011},
  publisher={Now Publishers, Inc.}
}

@article{pilanci2017newton,
  title={Newton sketch: A near linear-time optimization algorithm with linear-quadratic convergence},
  author={Pilanci, Mert and Wainwright, Martin J},
  journal={SIAM Journal on Optimization},
  volume={27},
  number={1},
  pages={205--245},
  year={2017},
  publisher={SIAM}
}

\appendix

\section{Related Work}
\label{apx:related}

Second-order methods in modern machine learning are often better understood as step pipelines than as monolithic algorithms.
A typical pipeline combines a curvature model
(e.g., Hessian, generalized Gauss-Newton, or Fisher),
an approximation or estimator
(e.g., diagonal probing, Kronecker factors, or axis-wise structure),
a linear solver
(direct or iterative; parameter-space or row-space),
optional preconditioning and warm starts,
and a damping policy.
Across method families, 
prior work shows that changing these components can materially affect stability and wall-clock time-to-accuracy.
This motivates systems abstractions that make those choices explicit and measurable.

\paragraph{Second-order methods in machine learning.}
Early work on second-order optimization for deep networks leveraged 
Hessian-free and truncated-Newton schemes implemented with 
conjugate gradients~\cite{martens2010deep,martens2011learning}.
A central lesson from subsequent studies is that practical behavior 
depends strongly on how the inner linear solve is 
composed with damping and step control, e.g., 
ablations of Hessian-free style training document sensitivity
to solver-damping interactions and other low-level design choices
that are typically omitted from the method description~\cite{wiesler2013investigations,martens2010deep,martens2011learning}.
In particular, studies of Hessian-free style training show that 
changing only the damping update signal or the inner-loop initialization 
(e.g., warm-starting CG from the previous solution) 
can change the amount of Krylov work substantially, 
with downstream effects on time-to-accuracy~\cite{wiesler2013investigations}.
This supports treating the solver-damping loop 
as a first-class design choice rather than a minor implementation detail.
Natural-gradient variants such as K-FAC similarly combine 
a specific curvature surrogate with 
structured approximations and non-trivial damping and scheduling 
choices~\cite{martens2015optimizing}.
Empirically, robust K-FAC behavior depends not only on the factorization itself, 
but also on how damping and rescaling are applied.
This reinforces that the curvature approximation 
and the control policy must be specified together to define the practical method~\cite{martens2015optimizing}.
Structure-aware preconditioners such as Shampoo exploit 
tensor axes to reduce cost~\cite{gupta2018shampoo}, 
while diagonal-curvature methods (e.g., AdaHessian, Sophia) 
estimate Hessian or Gauss-Newton diagonals and 
combine smoothing and clipping to keep overhead close to first-order 
training~\cite{yao2021adahessian,liu2023sophia}.
Recent Gauss-Newton variants revisit stochastic 
generalized Gauss-Newton operators solved by CG with autodiff-based 
Hessian-vector products (HVPs)~\cite{gargiani2020promise} 
as well as row-space realizations that change 
the cost model without changing the underlying subproblem~\cite{ren2019efficient,korbit2025exact}.
Taken together, 
these lines of work reinforce the compositional view:
solver choice, damping, curvature modeling, and estimation are coupled,
and the system should expose them as explicit axes for controlled ablation.

\paragraph{Scaling and systems.}
Systems work has pushed K-FAC and natural-gradient training to larger scales, emphasizing refresh policies, factor staleness, and communication patterns~\cite{osawa2019large,osawa2020scalable}.
Distributed implementations study how factor construction and inversion are distributed across layers, 
as well as decoupled updates and lower communication volume in all-reduce operations~\cite{pauloski2022deep,zhang2022scalable}.
More broadly, hybridization that mixes higher-order preconditioners 
with first-order updates in suitable bases 
(e.g., SOAP: Adam in Shampoo's eigenbasis) 
highlights practical trade-offs between preconditioning frequency, 
robustness, and cost~\cite{vyas2024soap}.
Together, these results support treating second-order training as an execution problem in which scheduling, reuse, 
and cost control are first-class concerns.

\paragraph{Curvature telemetry and Hessian spectra.}
Curvature measurements can be made online
at relatively low cost.
PyHessian, for example, provides scalable estimates of top eigenvalues, traces, and spectral densities, 
which can inform damping and step-size control~\cite{yao2020pyhessian}.
Large-scale studies of deep-network Hessian spectra analyze 
training dynamics and sample-size effects~\cite{papyan2018full}.
BackPACK shows how to expose per-example gradients 
and curvature quantities with minimal overhead in common 
frameworks~\cite{dangel2019backpack}, 
while Cockpit demonstrates the utility of such
dense telemetry for real-time debugging and tuning~\cite{schneider2021cockpit}.
For trace estimation, Hutch++ reduces probe complexity and makes frequent estimation more practical~\cite{meyer2021hutchpp}.

\paragraph{Libraries and implementations.}
Our abstraction builds on the Optax gradient-transformation 
chain design~\cite{deepmind2020jax} 
while targeting step-level semantics required by curvature-aware methods.
It complements solver-centric packages such as 
JAXopt~\cite{jaxopt_implicit_diff}
and Lineax~\cite{rader2023lineax}, which provide optimization and linear-solve primitives but do not manage the optimizer state required for curvature-aware training.
KFAC-JAX offers a modular JAX implementation of K-FAC and related 
estimators~\cite{kfac-jax2022github}.
Optimistix separates search and descent in JAX, 
providing a useful design precedent for decomposing optimizer logic~\cite{optimistix2024}.
Production-oriented implementations of Shampoo 
and axis-preconditioners in the JAX ecosystem further illustrate 
memory and communication trade-offs 
that motivate execution-aware system layers~\cite{duvvuri2024caspr}.

\paragraph{Benchmarking optimizers.}
Standardized evaluations caution that optimizer 
rankings are highly sensitive to tuning and protocol. 
Suites like DeepOBS, BenchOpt, and broad empirical studies 
advocate time-to-accuracy metrics and fair search 
spaces~\cite{schneider2019deepobs,schmidt2021descending,moreau2022benchopt}.
Our evaluation follows these recommendations by reporting steady-state step time and one-toggle ablations 
that isolate module and execution-lane choices under a fixed step contract.

\section{Usage and Reproducibility}
\label{apx:api}

This section documents the Somax API surface used in Section~4 
of the main paper.
Somax exposes a functional \texttt{init/step} interface 
backed by declarative specifications, allowing researchers to swap modules
(e.g., replacing \texttt{SolverSpec} from CG with Row-Cholesky)
without rewriting the training interface.
Code corresponding to this paper is available at \url{https://github.com/cor3bit/somax}.

\subsection{Core Interface}

The entry point \texttt{somax.assemble(spec, ...)} 
transforms a configuration into a \texttt{SecondOrderMethod} 
with a JAX-compatible signature.
For reproducibility, we also provide named presets
(e.g., \texttt{somax.make("newton\_cg")})
that recover the default hyperparameter settings used in this paper.

\begin{verbatim}
# 1. Construction
method = somax.make("newton_cg", curvature_kwargs={...}) 
# OR
method = somax.assemble(curvature=..., solver=..., damping=...)

# 2. Execution (JIT-compatible)
state = method.init(params)

@jax.jit
def train_step(params, batch, state, rng):
    # The step fuses linearization, solve, and updates
    params_next, state_next, info = method.step(params, batch, state, rng)
    return params_next, state_next, info
    
\end{verbatim}

\subsection{Declarative Specification System}

Somax separates the definition of a method from its execution plan.
Table~\ref{tab:api_specs} summarizes the primary specification objects 
used to compose the optimizers in the main paper (Experiments section).
These specs are data containers (Pytrees) that 
the planner analyzes to determine the execution lane and telemetry schema.

\begin{table}[t]
\centering
\caption{Primary configuration objects (Specs) in the Somax API.}
\label{tab:api_specs}
\small
\begin{tabular}{@{}l p{0.65\columnwidth}@{}}
\toprule
\textbf{Spec Object} & \textbf{Role \& Planner Effect} \\
\midrule
\texttt{CurvatureSpec} & Defines the operator family (e.g., \texttt{hessian}, \texttt{ggn\_ce}). Declares if row-primitives are available. \\
\texttt{SolverSpec} & Selects the algorithm (e.g., \texttt{cg}, \texttt{row\_cholesky}). \textbf{Determines the execution lane} (Param vs.\ Row). \\
\texttt{DampingSpec} & Defines the trust-region or regularization policy. Requests telemetry (e.g., rho-packs) that the planner must schedule. \\
\texttt{PrecondSpec} & (Optional) Defines diagonal scaling (e.g., \texttt{diag\_ema}) for the Diagonal lane or as a preconditioner for CG. \\
\texttt{EstimatorSpec} & (Optional) Layers randomized probing (e.g., Hutchinson) onto the step without extra linearizations. \\
\bottomrule
\end{tabular}
\end{table}

\subsection{Example: Assembling a Custom Optimizer}

The following example shows how to assemble a more complex method:
a Generalized Gauss-Newton (GGN) optimizer with trust-region damping,
EMA preconditioning, and a custom Optax chain.
The planner automatically resolves the dependency 
between the trust-region policy
(which requires \texttt{rho}) and the telemetry machinery
(which must schedule \texttt{loss\_after}).

\begin{verbatim}
import somax
import optax
from somax.specs import *

# 1. Define the Spec
method = somax.assemble(
    # Use Cross-Entropy GGN
    curvature=CurvatureSpec("ggn_ce", kwargs={"predict_fn": model.apply}),
    
    # Solve in Parameter Space using PCG
    solver=SolverSpec("cg", kwargs={"maxiter": 20, "tol": 1e-4}),
    
    # Precondition with EMA diagonal (Adam-style heuristic)
    precond=PrecondSpec("diag_ema", kwargs={"beta": 0.99}),
    
    # Use Trust-Region Damping
    damping=DampingSpec("trust_region", lam0=1.0),
    
    # Embed standard Optax transforms for the update application
    tx=optax.chain(
        optax.clip_by_global_norm(1.0),
        optax.scale(-1.0) # Descent direction
    )
)

# 2. The method is now a compiled JAX Pytree ready for .init/.step
\end{verbatim}

The assembler resolves defaults and validates compatibility at build time.
This is where Somax enforces lane-specific constraints, 
so the traced step contains no runtime compatibility 
checks beyond the lane fixed by the plan.
Table~\ref{tab:somax-contracts} summarizes the contracts used throughout the paper.

\begin{table}[tbp]
\centering
\caption{
\textbf{Planner-aware contracts (math $\rightarrow$ components).}
Each component declares requirements that the 
planner resolves into a static plan (lane, metric schema, cadences),
so the executor can fuse the step without redundant recomputation.
}
\label{tab:somax-contracts}
\setlength{\tabcolsep}{6pt}
\renewcommand{\arraystretch}{1.15}
\begin{tabularx}{\linewidth}{@{}p{0.14\linewidth} Y Y p{0.26\linewidth} p{0.16\linewidth}@{}}
\toprule
\textbf{Component} & \textbf{Consumes} & \textbf{Produces} & \textbf{Planner effect (why it matters)} & \textbf{Persistent state} \\
\midrule
Curvature operator
& parameters, batch
& linearization state; matrix-free primitives (and optionally row primitives); loss/grad consistent with snapshot
& determines feasibility of the row lane; declares whether post-update 
probes can reuse the step-local snapshot
& none (the linearization snapshot is step-local)
 \\[3pt]

Estimator wrapper (optional)
& curvature primitives and RNG (cadence-gated)
& additional curvature summaries (e.g., diagonal, trace, spectrum)
& adds optional probes without creating a second curvature snapshot; updates the fixed metric schema
& optional estimator state (e.g., probe buffers), else none \\[3pt]

Preconditioner (optional)
& diagonal estimate and damping
& diagonal scaling (used by diagonal lane or by PCG)
& controls whether the param lane specializes to CG vs PCG; avoids runtime switching inside traced code
& EMA buffers (optional) \\[3pt]

Linear solver
& lane-specific operator primitives, damping, right-hand side
& scaled direction; solver diagnostics (if enabled)
& fixes the lane (diag/param/row); enables cadence-gated solver telemetry keys
& warm start (solution to the previous iteration, optional) \\[3pt]

Damping policy
& plan-selected telemetry (e.g., gain-ratio bundle) at declared cadence
& updated damping state
& turns on post-update probes only when required; ensures probes are computed against the applied update
& damping state \\[3pt]

Optax post-transform
& scaled direction and Optax state
& applied update and updated Optax state
& forces step-level semantics: post-update probes must match the exact applied update, 
not the raw direction
& Optax state \\
\bottomrule
\end{tabularx}
\end{table}

\section{Experiment Details}
\label{apx:exp-details}

\subsection{Case Study I: Switching Lanes}
\label{apx:exp:a2}

This experiment corresponds to the lane-scaling study in
Section~4.2. 
Its purpose is to isolate how execution-lane choice 
changes wall-clock cost under a fixed step contract.

\paragraph{Dataset.}
We use a synthetic scalar regression task generated with
sklearn \texttt{make\_regression}.
The dataset contains $200{,}000$ examples with input dimension $128$ and additive noise $0.1$.
Data generation and the train/test split 
use a fixed seed so that all compared runs 
see the same underlying problem instance.
We use a $90/10$ train/test split.

\paragraph{Model.}
The model is a depth-$3$ ReLU MLP 
with scalar output.
For the width sweep, we vary the hidden width in
$\{32, 64, 128, 256, 512, 1024, 2048, 4096\}$
while keeping batch size fixed at $128$.
For the batch sweep, we fix hidden width at $128$ 
and vary batch size in
$\{32, 64, 128, 256, 512, 1024, 2048, 4096\}$.

\paragraph{Compared methods.}
We compare the Somax presets SGN~\cite{gargiani2020promise} \texttt{sgn\_mse} and EGN~\cite{korbit2025exact}
with CG solver \texttt{egn\_mse\_cg},
which instantiate the same outer training setup but execute in different lanes:
parameter space versus row space.
Both use the same learning rate $10^{-3}$,
constant damping with $\lambda_0 = 1.0$,
CG tolerance $10^{-5}$,
maximum $5$ CG iterations per step,
and warm-started inner solves.
Thus, execution lane is the only intended systems-level difference between the compared configurations.

\paragraph{Measurement protocol.}
All runs execute on a single NVIDIA RTX A4000 under JAX \texttt{jit}.
Before timing, 
we perform two warmup steps to exclude compilation 
and one-time setup from the reported step-time measurements.
Each run then executes for at most $2000$ optimization steps.
The script records wall-clock step time over 
windows of $100$ training steps.
For each run, it stores the mean, median, 
and standard deviation of these per-window timings.
The main figure uses \texttt{step\_time\_median\_ms} as the plotted metric.
With the current runner configuration, 
the study uses 5 seeds.
For each seed, 
we first compute a per-run median step time over timing windows.
Each plotted point is then the median of these five per-seed summaries.

\subsection{Case Study II: Module Interactions}
\label{apx:exp:a3}

This experiment corresponds to the module-interaction study in Section~4.3.
Its purpose is to isolate how solver modules interact within a fixed second-order execution regime.

\paragraph{Dataset.}
We use Fashion-MNIST, 
with the standard training and test splits.
Training data are loaded into host memory, 
while the full test set is placed on device for evaluation.
No data augmentation is used.

\paragraph{Model.}
The model is a medium-size 
convolutional neural network (CNN).
It consists of two convolution-ReLU-max-pooling stages
followed by three dense layers with ReLU activations 
and a final 10-class output layer.
All compared runs use the same model architecture and batch size $128$.

\paragraph{Compared methods.}
We include SGD and Adam as first-order baselines.
For second-order runs, 
we fix the solver family to parameter-space SGN.
Within this fixed lane, we vary three modules:
(i) damping policy: constant damping or trust-region damping;
(ii) CG budget: a light or heavy budget;
and (iii) diagonal preconditioning: on or off.
When enabled, the diagonal preconditioner 
is built from elementwise squared batch gradients.
The light budget uses $\texttt{cg\_maxiter}=3$, $\texttt{cg\_tol}=10^{-3}$, $\texttt{cg\_stabilise\_every}=10$, and warm start enabled.
The heavy budget uses $\texttt{cg\_maxiter}=10$, $\texttt{cg\_tol}=10^{-5}$, $\texttt{cg\_stabilise\_every}=5$, and warm start enabled.
For trust-region damping, we use $\texttt{tr\_every\_k}=5$, lower/upper thresholds $(0.25, 0.75)$, multiplicative update factors $(0.5, 1.5)$, and damping bounds $[10^{-12}, 10^6]$ with $\rho$ clipping at $5.0$.
All SGN runs use $\lambda_0 = 1.0$.

\paragraph{Training and measurement protocol.}
All runs use 10 epochs, batch size $128$, 
and execute on a single GPU under JAX \texttt{jit}.
Training data are iterated from precomputed 
epoch permutations so that epoch-boundary reshuffling does not contaminate timing windows.
Reported time-to-target denotes wall-clock time 
to reach 85\% test accuracy.

\paragraph{Learning-rate selection and seed aggregation.}
For each solver configuration, we perform a learning-rate search.
After selecting the learning rate for each configuration, 
the final reported table values are recomputed 
on the selected runs and aggregated over 5 seeds.

\subsection{Case Study III: Composability and New Optimizer Regimes}
\label{apx:exp:a5}

This experiment corresponds to the 
composability study in
Section~4.4.
Its purpose is to test whether Somax modularity 
can support not only named optimizer families, 
but also new optimizer regimes instantiated 
and evaluated under the same training loop.

\paragraph{Dataset.}
We use CIFAR-10 with the standard training and test splits.
Training examples are loaded from TensorFlow Datasets 
and normalized channel-wise using 
the standard CIFAR-10 mean and standard deviation.
For training, we apply standard data augmentation consisting 
of zero-padding to $40 \times 40$, 
random cropping back to $32 \times 32$, 
and random horizontal flipping.
Test data are normalized but not augmented.

\paragraph{Model.}
The model is a CIFAR-style ResNet-20.
It uses a $3 \times 3$ convolutional stem with 16 channels, 
followed by three residual stages with channel widths 16, 32, and 64.
Each stage contains 3 basic residual blocks, 
yielding the standard ResNet-20 depth for CIFAR.
Batch normalization is used throughout, 
with global average pooling and a final dense classification layer.
All compared runs use batch size $128$.

\paragraph{Compared methods.}
We compare SGDM, Sophia-H~\cite{liu2023sophia}, 
Sophia-G~\cite{liu2023sophia}, 
and a modularly recombined variant denoted Sophia-N.
SGDM is included as a first-order baseline.
The three Sophia-style methods share the same outer update 
but differ in how curvature information is constructed.
Sophia-H uses Hutchinson-style diagonal estimation
based on the Exact Hessian curvature operator.
Sophia-G uses a Gauss-Newton-Bartlett-style diagonal estimate based on sampled targets.
Sophia-N combines GGN curvature with Hutchinson diagonal estimation.
This yields a new Sophia-style configuration obtained through modular recombination in Somax.
Sophia-G uses \texttt{gamma}=0.05, \texttt{n\_samples}=1, and \texttt{eval\_every\_k}=10.
Sophia-H uses \texttt{gamma}=0.01, \texttt{n\_probes}=1, and \texttt{eval\_every\_k}=10.
Sophia-N uses \texttt{gamma}=0.05, \texttt{n\_probes}=1, and \texttt{eval\_every\_k}=10.

\paragraph{Training and measurement protocol.}
All runs use 200 epochs, batch size $128$, 
and execute on a single GPU under JAX \texttt{jit}.
SGDM uses momentum $0.9$, weight decay $5 \times 10^{-4}$, 
and a step learning-rate schedule.
All Sophia-style runs use cosine learning-rate decay 
with 2000 warm-up steps.
Test evaluation is performed every 500 training steps.

\paragraph{Learning-rate selection and seed aggregation.}
For each solver configuration, 
we perform a learning-rate search.
After selecting the learning rate and 
learning-rate schedule 
for each configuration, 
the final reported table values are recomputed 
on the selected runs and aggregated over 5 seeds.

\end{document}